\documentclass[10pt, conference]{IEEEtran}

\usepackage[
letterpaper,
left=0.64in,
right=0.63in,
top=0.75in, 
bottom=1.05in]{geometry}

\usepackage[utf8]{inputenc}
\usepackage{graphicx} 
\usepackage{caption}
\usepackage{subcaption}
\usepackage{amsmath,amssymb,amsfonts}
\usepackage[ruled,linesnumbered]{algorithm2e}
\usepackage{cite}
\usepackage{xcolor}
\usepackage{soul}
\usepackage{tikz}
\usetikzlibrary{fit,calc}
\usepackage{balance}

\usepackage{booktabs} 

\newcommand{\etal}{\textit{et al}. }

\newcommand*{\tikzmk}[1]{\tikz[remember picture,overlay,] \node (#1) {};\ignorespaces}
\newcommand{\boxit}[1]{\tikz[remember picture,overlay]{\node[yshift=3pt,fill=#1,opacity=.25,fit={(A)($(B)+(.8\columnwidth,.8\baselineskip)$)}] {};}\ignorespaces}
\colorlet{mypink}{red!40}

\IEEEoverridecommandlockouts
\IEEEpubid{\begin{minipage}{\textwidth}\centering
© 2025 IEEE. Personal use of this material is permitted. Permission from IEEE must be obtained for all other uses, in any current or future media, including reprinting/republishing this material for advertising or promotional purposes, creating new collective works, for resale or redistribution to servers or lists, or reuse of any copyrighted component of this work in other works. https://www.ieee.org/publications/rights/rights-policies.html
\end{minipage}}

\title{Federated Learning for Anomaly Detection in Energy Consumption Data: Assessing the Vulnerability to Adversarial Attacks}

\author{
    \IEEEauthorblockN{
        Yohannis Kifle Telila\IEEEauthorrefmark{1}, 
        Damitha Senevirathne\IEEEauthorrefmark{1},
        Dumindu Tissera\IEEEauthorrefmark{1},}
    \IEEEauthorblockN{
        Apurva Narayan\IEEEauthorrefmark{1}\IEEEauthorrefmark{2},
        Miriam A.M. Capretz\IEEEauthorrefmark{1},
        and Katarina Grolinger\IEEEauthorrefmark{1}}
    \IEEEauthorblockA{
        \IEEEauthorrefmark{1} \textit{Department of Electrical and Computer Engineering}, 
        \IEEEauthorrefmark{2} \textit{Department of Computer Science} \\
        \textit{Western University,}
        London, Ontario, Canada
    }
}

\date{June 2024}

\begin{document}
\bstctlcite{IEEEexample:BSTcontrol}
\maketitle
\IEEEpubidadjcol

\begin{abstract}
Anomaly detection is crucial in the energy sector to identify irregular patterns indicating equipment failures, energy theft, or other issues. Machine learning techniques for anomaly detection have achieved great success, but are typically centralized, involving sharing local data with a central server which raises privacy and security concerns. Federated Learning (FL) has been gaining popularity as it enables distributed learning without sharing local data. However, FL depends on neural networks, which are vulnerable to adversarial attacks that manipulate data, leading models to make erroneous predictions. While adversarial attacks have been explored in the image domain, they remain largely unexplored in time series problems, especially in the energy domain. Moreover, the effect of adversarial attacks in the FL setting is also mostly unknown. This paper assesses the vulnerability of FL-based anomaly detection in energy data to adversarial attacks. Specifically, two state-of-the-art models, Long Short Term Memory (LSTM) and Transformers, are used to detect anomalies in an FL setting, and two white-box attack methods, Fast Gradient Sign Method (FGSM) and Projected Gradient Descent (PGD), are employed to perturb the data. The results show that FL is more sensitive to PGD attacks than to FGSM attacks, attributed to PGD's iterative nature, resulting in an accuracy drop of over 10\% even with naive, weaker attacks. Moreover, FL is more affected by these attacks than centralized learning, highlighting the need for defense mechanisms in FL.
\end{abstract}


\section{Introduction}
\label{sec:introdcution}
\vspace{-0.02in}
In 2022, commercial and residential buildings accounted for approximately 34\% of global energy consumption while contributing to 37\% of energy and process-related carbon dioxide (CO2) emissions \cite{UNEP2024}. The United Nations' Net-zero emission goal aims to reduce the adverse effects of global warming by decreasing global net CO2 emissions to 55\% of 2010 levels by 2030, with the target of achieving net zero by 2050 \cite{UNEP2024Net}. Therefore, identifying and mitigating energy waste is essential to align with the net-zero emissions goal. Anomaly detection, the process of identifying patterns that diverge from the established normal behavior \cite{AnomalyChandola}, plays a crucial role in determining energy waste. Example causes of anomalies in energy consumption data include wasteful human usage \cite{pan2022high}, faulty control systems \cite{elmasry2022enhanced}, and energy theft \cite{yip2018anomaly}. These anomalies can be detected by identifying deviations from the consumer's historical energy patterns, which can then trigger energy-conserving measures, promoting efficiency and reducing environmental impact.

\IEEEpubidadjcol  
\IEEEpubidadjcol

Energy distribution companies have been transitioning to smart meters, which measure and transmit energy consumption information for improved energy management and control \cite{yip2018anomaly}. The potential of smart meter data for anomaly detection has been widely recognized, and several techniques have been proposed, with Machine Learning (ML) approaches providing state-of-the-art solutions. Common approaches to training ML models involve transmitting data to a central server for model training on that server \cite{jithish2023distributed}. However, this centralized approach exposes data to privacy and security risks.

Federated Learning (FL) \cite{kairouz2021advances} reduces these privacy and security risks by decentralizing the learning with multiple nodes collaboratively training a global model without sharing their local data. Upon receiving the global model weights from the server during an FL round, clients train independently on their local data for several iterations and subsequently send the updated model weights to the server. The server aggregates clients' weights to update the global model, which is then broadcast back to the clients for the next training round. By keeping data local, FL enhances privacy and security while facilitating compliance with regulatory requirements such as the EU/UK General Data Protection Regulation \cite{TRUONG2021102402}. Recognizing these advantages, energy consumption studies have integrated FL to train neural networks and preserve data locality \cite{DistributedLoad2022, AsynchronousFL}. Thus, our study focuses on FL-based anomaly detection in smart meter energy consumption data.

While FL mitigates privacy and security risks associated with data sharing, it still raises other security concerns \cite{boenisch2023curious, fowl2021robbing, Bouacida2021} such as the vulnerability to adversarial attacks from malicious clients \cite{Bouacida2021}. Several studies \cite{zhang2023delving, chen2022gear} in vision tasks showed that FL is vulnerable to white-box adversarial attacks targeting client neural networks (NNs), such as the Fast Gradient Sign Method (FGSM) \cite{goodfellow2014explaining} and Projected Gradient Descent (PGD) \cite{madry2017towards}. These attacks effectively degrade NN performance by exploiting the network's vulnerabilities through subtle perturbations of input data. Initially designed for the computer vision domain, these attacks have subsequently been adopted in time series tasks \cite{fawaz2019adversarial, mode2020adversarial}, demonstrating their transferability. However, in the time series domain, these attacks still focus on centralized ML and do not examine the effect of attacks in the FL setting. As NNs are deployed on client nodes in FL, the attack surface expands, increasing the risk as individual compromised nodes can affect the complete federation. Therefore, it is crucial to understand how adversarial attacks affect the FL process.

Consequently, this study assesses the vulnerability of FL-based anomaly detection in energy consumption to adversarial attacks. Specifically, the study examines how attacks on individual FL nodes affect the global model and impact overall FL-based anomaly detection performance, focusing on the effects of different adversarial attacks. We adopt state-of-the-art time series neural networks, Long-Short Term Memory (LSTM) and Transformer, given their proven effectiveness in the energy consumption domain \cite{DistributedLoad2022, wang2020lstm, l2022transformer, zhang2021power}. The impact of attack magnitude in both non-iterative FGSM and iterative PGD attack methods within the FL setting is examined. Further, the FGSM and PGD attack methods are compared against simpler techniques: label flipping \cite{shen2023privacy} and random perturbation. Finally, we evaluate the impact of the number of malicious clients in the FL environment. The results indicate the vulnerability of FL-based anomaly detection in energy consumption to adversarial attacks, regardless of the underlying model, highlighting the need for further research on resilient FL techniques.

The remainder of this paper is organized as follows: Section \ref{sec:relatedwork} discusses related work, Section \ref{sec:background} provides background, Section \ref{sec:methodology} describes the methodology, and Section \ref{sec:experiments} presents the result and finding. Finally, Section \ref{sec:conclusion} concludes the paper.

\section{Related Work}
\label{sec:relatedwork}
\vspace{-0.02in}
This section reviews energy anomaly detection in FL and discussed adversarial attacks.

\subsection{Energy Anomaly Detection and Federated Learning}
\label{sub_sec:AnomalyFL}

Anomaly detection in energy has been an active research area due to its critical importance in detecting sub-optimal performance, device malfunction, and abnormal behavior, thereby contributing to energy efficiency. Several studies  \cite{AnomalyDNN, ChalaDNN2020, somu2021deep, srCNN2021} proposed anomaly detection based on Deep Neural Networks (DNN) due to their ability to model complex relationships. Recently, LSTMs have gained popularity in the energy domain due to their ability to capture temporal dependencies \cite{ChalaDNN2020, somu2021deep}. 

In the Natural Language Processing (NLP) domain, Transformer has been renowned for its attention mechanism and the generative capabilities utilized in diverse applications such as ChatGPT \cite{achiam2023gpt}. Consequently, Transformer has been adapted in various energy data studies \cite{nazir2023forecasting, zhang2021power}. Energy forecasting is commonly used with Transformers in self-supervised anomaly detection solutions. Such solutions examine the difference between the actual and predicted values: if this difference exceeds a threshold, the sample is deemed abnormal. Zhang \etal \cite{zhang2021power} combined Transformer and K-means clustering to forecast energy consumption and detect anomalies. Nazir \etal \cite{nazir2023forecasting} also presented a Transformer-based solution focusing on energy forecasting. As LSTMs and Transformers have dominated energy anomaly detection and forecasting, our study considered these two architectures in the FL setting. 

These energy forecasting/anomaly detection techniques typically train models centrally where the data sharing raises privacy and security risks. Due to its data privacy-enhancing capabilities and distributed nature, FL has been gaining popularity for various tasks in the energy domain. Fekri \etal \cite{DistributedLoad2022} proposed a distributed load forecasting method based on FL which takes advantage of LSTM as the base learner. The same group further advanced FL-based forecasting to enable asynchronous learning in the presence of non-IDD data by introducing a novel aggregation technique \cite{AsynchronousFL}. Similarly, Sater \etal \cite{Sater2020AFL} also integrated LSTM but combined it with multi-task learning for anomaly detection in smart buildings. The approach proposed by Jithish \etal \cite{jithish2023distributed} for anomaly detection in smart grids is also based on FL. They considered the diversity of ML models as the base learning, including LSTM, Gated Recurrent Neural Networks (GRU), and Vanilla RNN. 

The studies employing FL in the energy field made great strides in improving energy prediction and anomaly detection by enabling model training without sharing raw data. Nevertheless, they did not consider the possible presence of malicious clients in the federation. Consequently, we address this gap by examining the vulnerability of FL-based anomaly detection techniques to adversarial attacks.

\subsection{Adversarial Attacks}
\label{sub_sec:AdversarialAttack}

Adversarial attacks are designed to deceive the ML model, leading to incorrect classifications. Often examined in computer vision, adversarial samples are created by altering the images so they still look normal to the human eye but lead the model to incorrect predictions. Goodfellow \etal \cite{goodfellow2014explaining} proposed the Fast Gradient Sign Method (FGSM) that utilizes the gradients of a neural network to craft adversarial input image samples. In the FGSM approach, the gradient is calculated only once. To extend this, Kurakin \etal \cite{kurakin2017adversarial} introduced the Basic Iterative Method (BIM), an iterative approach that adds perturbations to the input data. Similarly, Madry \etal \cite{madry2017towards} introduced another iterative attack named Projected Gradient Descent (PGD). PGD uses random initialization and a projection step which iteratively alters the input to improve the generated samples. However, these algorithms were designed for the computer vision domain, adding perturbations to images.

Considering the transferability of the mentioned attack models, Fawaz \etal \cite{fawaz2019adversarial} emphasized that adversarial attacks have not been thoroughly explored for time series classification. Thus, they perturbed time series data using FGSM and BIM and compared the effectiveness of the attack models. Mode \etal \cite{mode2020adversarial} also considered adversarial attacks on multivariate time-series data and adapted adversarial attacks from the image domain, such as FGSM and BIM, to deep learning regression models for multivariate time series forecasting. However, these studies  \cite{fawaz2019adversarial, mode2020adversarial} only considered centralized ML.  

Overall, adversarial attacks were mostly considered in the vision domain, with a few works investigating time-series \cite{fawaz2019adversarial, mode2020adversarial} but in the centralized setting. Furthermore, Bondok \etal \cite{bondok2023novel} studied attacks in an FL setting; however, they focused on theft detection while our study focuses on generic anomaly detection. Nevertheless, there is a need to understand how adversarial attacks affect anomaly detection in the FL setting. To address this gap, our work investigates the vulnerability of FL-based anomaly detection models, specifically LSTM and Transformer, to adversarial attacks of various strengths and compares their vulnerability to that of centralized training.

\section{Background}
\label{sec:background}
\vspace{-0.02in}
This section provides an overview of the two white-box attacks adopted in this paper. In white-box attacks, the adversaries have full access to the trained model, including the model structure, parameters, and training data. White-box attacks were selected over black-box attacks because access to the model's architecture and parameters enables more effective adversarial attacks. Specifically, the two white-box attacks considered, FGSM and PGD, perturb the input sample in the direction of the model's gradient with respect to the input.

The FGSM, as depicted in Algorithm \ref{alg:fgsm_attack}, generates an adversarial sample by computing the gradient of the loss with respect to the input and adding a proportion of this gradient to the input as the perturbation. Given a model parameterized by $\theta$, an original input sample $x$, and its label $y$, the computation of the adversarial sample $x_{adv}$ by FGSM can be expressed as:
\begin{equation}
\label{eq:fgsm}
\text{x}_{adv} = x + \epsilon \cdot \text{sign}(\nabla_x J(\theta, x, y)),
\vspace{-0.04in}
\end{equation}
where \(\nabla_x J(\theta, x, y)\) represents the gradient of the loss function \(J\) with respect to input $x$, and \(\text{sign}(\cdot)\) function extracts the sign of each element in the gradient. A scalar \(\epsilon\) controls the magnitude of the perturbation.

\begin{algorithm}[b]
\caption{FGSM attack algorithm}
\label{alg:fgsm_attack}
\SetKwInOut{Input}{Input}
\SetKwInOut{Output}{Output}

\Input{Neural network model \( M \), input data \( x \), true labels \( y \), magnitude of the perturbation \( \epsilon \)}
\Output{Adversarial data \( x_{adv} \)}

\SetKwFunction{FMain}{FGSM\_Attack}
\SetKwProg{Fn}{Function}{:}{}
\Fn{\FMain{\( M, x, y, \epsilon \)}}{
    Calculate \(\nabla_x J(\theta, x, y)\), gradient of the loss w.r.t. input \( x \)\;
    Perform FGSM attack on input data \( x \): \( x_{adv} \leftarrow x + \epsilon \cdot \text{sign}(\nabla_x J(\theta, x, y)) \)\;
}
\end{algorithm}

While FGSM is a one-step attack, the PGD is a multi-step variant of the FGSM, offering increased effectiveness over FGSM. As illustrated in Algorithm \ref{alg:pgd_attack}, the iterative loop of the PGD algorithm perturbs the input sample incrementally in multiple sequential steps. Given a model parameterized with $\theta$, an original input sample $x$ and its label $y$, in the $t^{th}$ iteration, PGD updates the adversarial sample from previous step $x_{adv}^{t-1}$ to the new adversarial sample $x_{adv}^{t}$ as follows:
\begin{equation}
\label{eq:pgd}
x_{adv}^t =  \left(x_{adv}^{t-1} + \epsilon \cdot \text{sign}(\nabla_x J(\theta, x_{adv}^{t-1}, y))\right).
\vspace{-0.02in}
\end{equation}
Here \(\nabla_x J(\theta, x_{adv}^{t-1}, y)\) is the gradient of the loss function with respect to the adversarial sample from the previous step, \(\text{sign}(\cdot)\) function extracts the signs of the gradient components, and \(\epsilon\) is the step size of the perturbation. 

\begin{algorithm}[b]
\caption{PGD attack algorithm}
\label{alg:pgd_attack}
\SetKwInOut{Input}{Input}
\SetKwInOut{Output}{Output}

\Input{Neural network model \( M \), input data \( x \), true labels \( y \), number of PGD iterations \( T \), PGD step size \( \epsilon \)}
\Output{Adversarial data \( x_{adv}^T \)}

\SetKwFunction{FMain}{PGD\_Attack}
\SetKwProg{Fn}{Function}{:}{}
\Fn{\FMain{\( M, x, y, T, \epsilon \)}}{
    Initialize \(\ x_{adv}^0 \)\ : \( \ x_{adv}^0  \leftarrow \ x \)\;
    \For{$t \leftarrow 1$ \KwTo $T$} {
        Calculate \(\nabla_x J(\theta, x_{adv}^{t-1}, y)\), gradient of the loss w.r.t. perturbed input \( x_{adv}^{t-1} \)\;
        Update perturbation: \(x_{adv}^t \leftarrow x_{adv}^{t-1} + \epsilon \cdot \text{sign}(\nabla_x J(\theta, x_{adv}^{t-1}, y)) \)\;
    }
}
\end{algorithm}

Both FGSM and PGD were initially proposed and subsequently examined using image data. Their consideration with time-series data has been limited \cite{fawaz2019adversarial}. In this study, we adapt them to generate attacks on anomaly detection in energy data.

\section{Methodology}
\label{sec:methodology}
\vspace{-0.02in}
This section outlines our methodology to assess the vulnerability of anomaly detection in energy data within the FL setting. First, the anomaly detection dataset generation is described, followed by the adaptation of the white-box attacks to perturb energy data. Then, the integration of adversarial attacks into the FL process by simulating adversarial clients is described. 

\subsection{Energy Anomaly Detection Data}
\label{sub_sec:AnomalyDataset}

The anomaly detection process requires transforming the energy data recorded by smart meters into a suitable input format for the anomaly detection model, in our case, LSTM or Transformer. The smart meter data, in the form of hourly energy consumption readings, are divided into segments representing 24-step daily load profiles. One input example for the anomaly detection model, thus, represents the daily energy consumption pattern over 24 hours. The resulting dataset of time-series energy consumption consists of the sets of inputs $\mathbf{X}$ and the corresponding labels $\mathbf{y}$:
\begin{multline}
\mathbf{X} = \{\mathbf{x}_1, \ldots,\mathbf{x}_i, \ldots,  \mathbf{x}_n \}, \quad \mathbf{y} = \{ y_1, \ldots, y_i, \ldots, y_n \} \\ where \quad 
\mathbf{x}_i = \begin{bmatrix}
x_{i1} & x_{i2} & \cdots & x_{i24}
\end{bmatrix}^T
\vspace{-0.02in}
\end{multline}
Here, input $\mathbf{x}_i$ is a load profile vector of day $i$ with 24 hourly readings, and ${y}_i$ indicates whether $\mathbf{x}_i$ is anomalous or not. 

While energy consumption data are abundant, labeled data are not commonly available, and this is also true for the dataset used in our study. As the anomalies represent rare events, we can assume that most recorded data are non-anomalous. Thus, we label all original load profiles as normal data. Consequently, to address the lack of anomalous data, we generate synthetic anomalies that mimic irregularities in energy consumption patterns. To generate synthetic anomaly data samples, the historical 24-hour energy usage data is analyzed first to identify periods of high and low electricity consumption (Referred to as high/low energy usage periods). 

Aligning with the energy consumption patterns, five distinct types of anomalies are introduced into the data. The first anomaly type is a \textit{drop}, which simulates a sudden and unexpected decrease in energy usage to zero, potentially caused by factors such as power outages or equipment failures. Given an input sample $\mathbf{x}_i$, and a randomly selected step $S$ from the high energy usage period, a length $l$ \textit{drop} anomaly is generated by modifying the time series entries $x_{iS}$ to $x_{i(S+l-1)}$ to zero:
\begin{equation}
\hat{x}_{ij | j = S \ldots S+l-1} = 0, \quad l \in \{1, 2\}.
\vspace{-0.05in}
\end{equation}

Additionally, \textit{positive spikes} and \textit{negative spikes} are introduced, representing increases or decreases in energy usage, respectively. These anomalies are introduced for a single time step (\( l=1 \)). Furthermore, \textit{segment positive spikes} and \textit{segment negative spikes} are introduced to represent deviations that persist for two timesteps (\(l=2\)). These anomalies could be indicative of events such as appliance malfunctions, unusual weather conditions, or changes in usage patterns. While in our experiment, only duration of \(l=2\) was considered, longer anomalies could be simulated. To generate spike anomalies, a random variable \( r \), sampled from the range \([0.5, 1.5]\), determines the amplitude of each spike or dip. For positive spike and segment positive spike anomalies, the modified value is:
\begin{equation}
\hat{x}_{ij | j = S \ldots S+l-1} = x_{ij} + r \cdot x_{ij} , \quad  l \in \{1, 2\},
\vspace{-0.05in}
\end{equation} 
where $S$ is from \textit{low} energy usage period. For negative spike and segment negative spike anomalies, the modified value is:
\begin{equation}
\hat{x}_{ij | j = S \ldots S+l-1} = x_{ij} - r \cdot x_{ij} , \quad  l \in \{1, 2\},
\vspace{-0.05in}
\end{equation}
where $S$ is chosen from \textit{high} energy usage period.

The resulting anomalous data points are combined with the original data to build the anomaly detection dataset. This study considers two models for anomaly detection: LSTM and Transformer. LSTM was selected for its effectiveness in capturing temporal patterns and its widespread use in energy modeling \cite{DistributedLoad2022}. Transformer was chosen due to its success in sequential tasks like GPT \cite{achiam2023gpt} and energy forecasting \cite{l2022transformer}.

\subsection{Time Series Adversarial Attacks}
\label{sub_sec:AttackingAnomalyData}

To create attacks on $\mathbf{X}$, the input data $\mathbf{x}_i$ needs to be perturbed to cause the model to misclassify it as $\hat{y_i}$ where $\hat{y_i}\neq y_i$. The FGSM attack is performed as in Equation~\ref{eq:fgsm}. While the input $x$ in Equation~\ref{eq:fgsm} is an image in the image domain, in our case, it is the daily load profile $\mathbf{x}_i$.

The gradients $\nabla_x J(\theta, x, y))$ calculated in Equation~\ref{eq:fgsm} depend on the model parameters $\theta$. The LSTM and Transformer models used for the anomaly detection contain different parameters $\theta$. Since the perturbation depends on the gradient of the loss, it will differ for each of the two models, potentially resulting in varying vulnerability to attacks between the two models. 

As for FGSM and PGD attacks, the gradients are calculated the same way. The difference, as illustrated in Algorithm \ref{alg:pgd_attack}, is that for PGD, the process is iterative. For both attacks, we examine the vulnerability of ML models under different attack strengths by varying parameter $\epsilon$.

\subsection{Adversarial Attacks in FL Setting}
\label{sub_sec:AttackingFL}

\begin{figure}[t]
    \centering
    \begin{subfigure}[b]{0.49\linewidth}
        \centering
        \includegraphics[width=\textwidth]{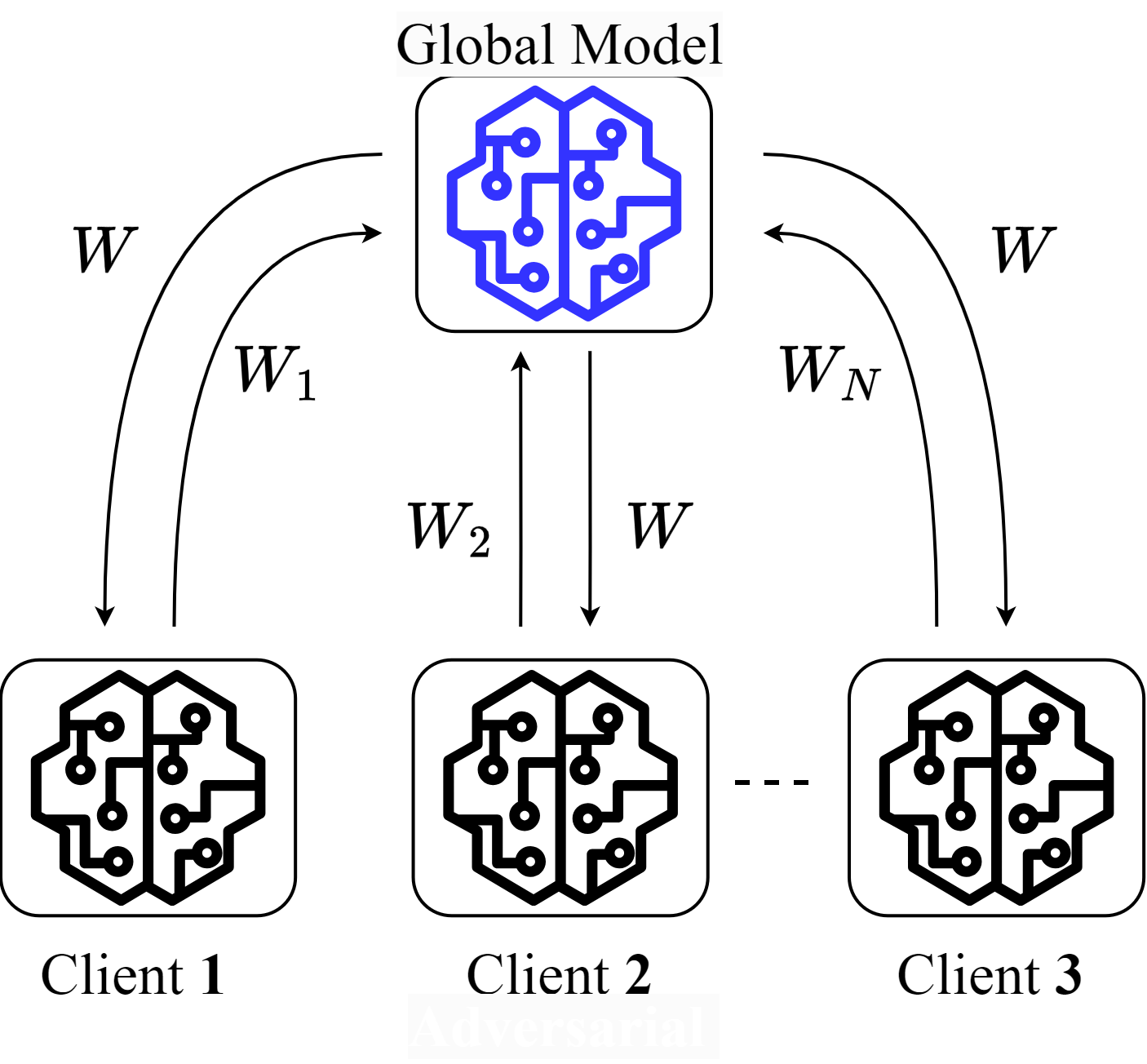}
        \caption{FL without adversarial attacks}
        \label{fig:fl_process}
    \end{subfigure}
    \hfill
    \begin{subfigure}[b]{0.49\linewidth}
        \centering
        \includegraphics[width=\textwidth]{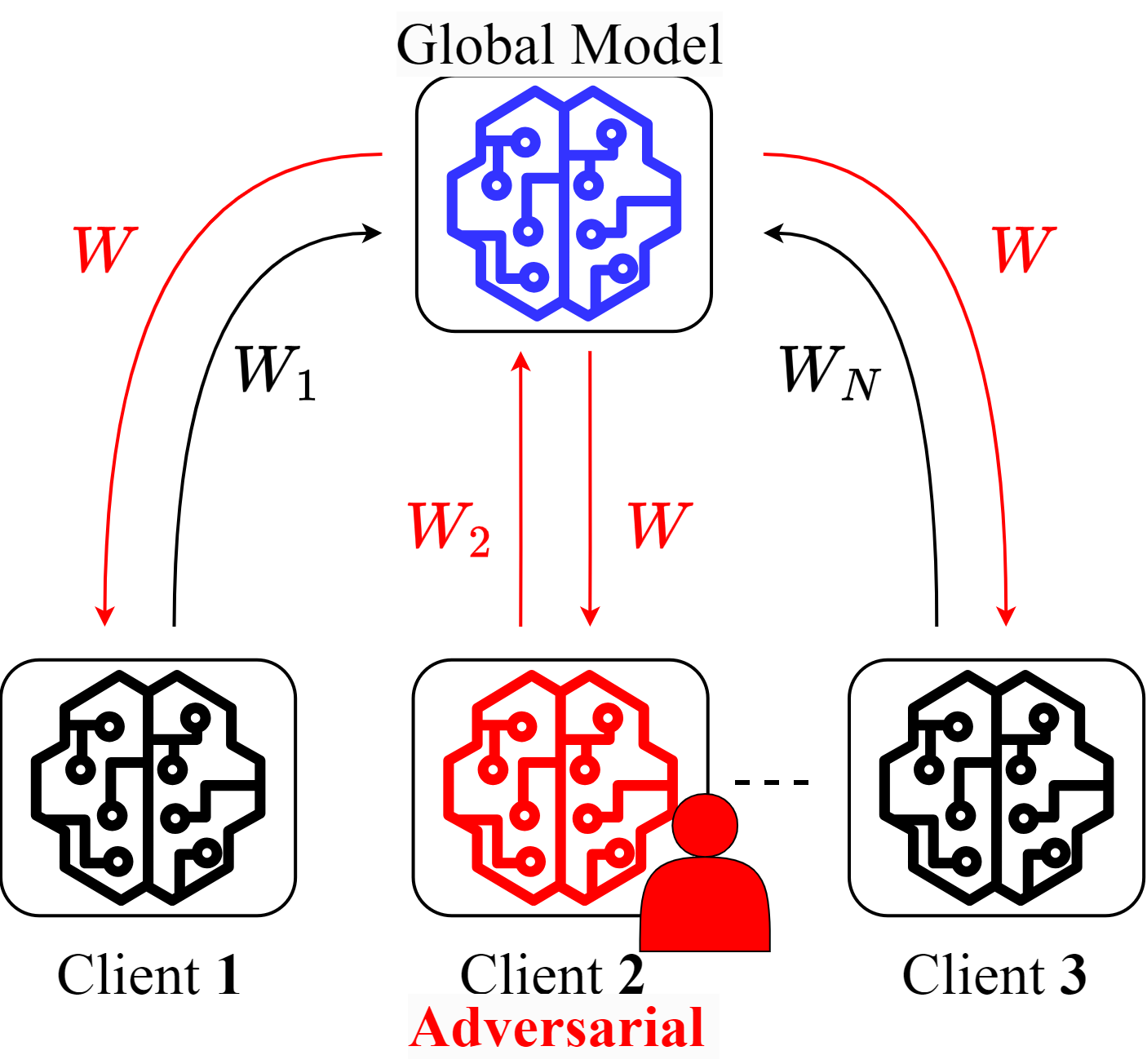}
        \caption{FL with adversarial attacks}
        \label{fig:fl_attack}
    \end{subfigure}
    \caption{Federated learning without and with adversarial attacks.}
    \label{fig:fl_combined}
\end{figure}

Consider an FL environment as in Fig. \ref{fig:fl_process} with a set $C$ of clients, i.e., houses with their energy consumption. The FL process with the integrated attack simulation is presented in Algorithm \ref{alg:fed_attack}. The highlighted lines 7 to 13 are added or modified to integrate the adversarial attacks into the FL process, while the remaining lines are traditional FL process. The training begins by randomly initiating global model weights, Line 1. The FL process iterates over $T$ rounds, Line 2, where each round starts with the global weights being broadcasted to a randomly selected subset $K$ of $N$ clients, Lines 3 and 4. 

The selected clients participate in the training round where they learn in isolation with their respective local datasets for $e$ number of epochs and update their local model weights $W_k$ -- without adversarial attacks, Line 12 in Algorithm \ref{alg:fed_attack}. Afterward, each client sends the locally updated weights to the global server for weight aggregation, Line 16. We consider the Federated Averaging (FedAVG) algorithm \cite{mcmahan2017communication} as a greatly popular aggregation technique. FedAVG averages client weights, Line 17, to compute the new global weights $W$ as follows: 
\begin{equation}
\label{eq:federated_average}
{W} = \frac{1}{N} \sum_{k \in K} W_k
\vspace{-0.05in}
\end{equation}
The process then continues to the next training round from Line 2 by selecting a new subset $K$ of $N$ clients to participate in training and broadcasting the new global weights $W$. The process repeats for $T$ training rounds or until convergence. 

With adversarial attacks, the set $C$ of all clients now contains a subset $A$ of adversarial clients whose training data are perturbed by adversarial attacks. As shown in Fig. \ref{fig:fl_attack}, if a particular adversarial client participates in the training round, its data is modified through FGSM/PGD attacks. Suppose adversarial clients are present among the selected subset $K$ for a particular training round. In that case, their data is perturbed by performing FGSM/PGD attacks, Line 9, with gradients calculated on the local models obtained in Line 8. The FGSM and PGD attacks are described in Section \ref{sec:background}.

The malicious clients in set $K$ train on perturbed data, Line 10, while non-malicious clients train on their original data, Line 12. Both types of clients send data to the server for aggregation, Lines 16 and 17. The presence of such perturbed local weights in the weight aggregation process results in subsequently broadcasting the effect of perturbed weights to all participating clients. Depending on the strength of the attack, modified weights $W_k$ can have a major impact on the global model aggregated by FedAVG. Therefore, malicious clients degrade the performance of the complete federation.

\begin{algorithm}[t]
\caption{Perform adversarial attacks in FL setting}
\label{alg:fed_attack}
\SetKwInOut{Input}{Input}
\SetKwInOut{Output}{Output}

\Input{Set of FL clients \(\{C\}\), number of training rounds \(T\), number of randomly selected clients per each training round \(N\), number of epochs to train a local model \(e\), randomly selected adversarial clients set \(\{A\}\)}
\Output{Global model weights \(W_{\text{global}}\)}
\BlankLine

Initialize \(W_{\text{global}}\) at the server\;
\For{$t \leftarrow 1$ \KwTo $T$} {
    Select set \(\{K\}\) using \(N\) random clients from \(\{C\}\)\;
    Server broadcasts \(W_{\text{global}}\) to all \(k \in \{K\}\) clients\;
    \ForEach{\(k \in \{K\}\) in parallel} {
        Client \(k\) receives \(W_{\text{global}}\)\;
        \tikzmk{A}
        \eIf{\(k \in \{A\}\)} {
            Get client model: \(M_k\)\;
            Perform attacks on client data \(x_k\): \(x_{adv} \leftarrow \text{Perform\_Attack}(M_k, x_k, y)\)\;
            Train local model on adversarial data: \(W_k \leftarrow \text{train}(W_k, x_{adv}, e)\)\;
        } {
            Train local model on non-adversarial data: \(W_k \leftarrow \text{train}(W_k, x_k, e)\)\;
        }
        \tikzmk{B}
        \boxit{mypink}
    }
    Global server aggregates \(W_k\) from all \(k \in \{K\}\) clients\;
    \(W_{\text{global}} \leftarrow \frac{1}{N} \sum_{k \in K} W_k\)\;
}
Server sends updated \(W_{\text{global}}\) back to all clients\;
\end{algorithm}

\section{Evaluation}
\label{sec:experiments}
\vspace{-0.02in}
This section describes the dataset and experimental setup, model training and evaluation metrics, followed by the effect of attacks during inference and training. Finally, the impact of the attack magnitude is examined.

\subsection{Dataset and Experimental Setup}
\label{sub_sec:DatasetandExperiment}

The empirical evaluation was conducted using a real-world residential energy consumption dataset provided by London Hydro, a local electricity distribution company serving the city of London, Ontario. The dataset comprises energy consumption data from 19 households, with each consumer's dataset containing hourly energy consumption readings for three years, resulting in 25,560 samples per household. The simulated FL setting treated each household with its associated energy consumption data as a distinct local node or client. 

\begin{figure}[t]
    \hspace{-0.3cm} 
    \includegraphics[trim={0 0.3cm 0 1cm},clip, width=0.95\columnwidth]{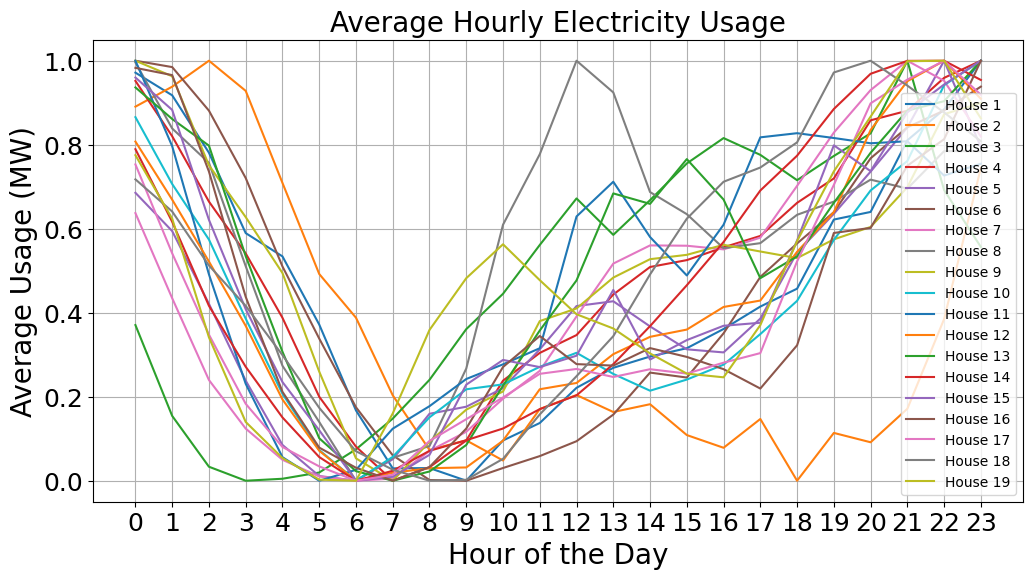}
    \caption{Average 24-hour electricity usage for each house.}
    \label{fig:ave-usage-pattern}
\end{figure}

The data were first segmented into daily load profiles as described in Section \ref{sub_sec:AnomalyDataset} and daily energy consumption patterns were analyzed. Fig. \ref{fig:ave-usage-pattern} represents the household-wise average 24-hour electricity usage pattern. The \textit{low} energy usage occurs between 4 a.m. and 10 a.m., and the \textit{high} energy usage from 6 p.m. to 1 a.m. These low and high periods were used to generate synthetic anomalous profiles and build the energy consumption anomaly detection dataset. The normal data were labeled 0, whereas synthetically generated anomalies were labeled 1.

\begin{figure}[t]
    \centering
\begin{subfigure}{0.48\columnwidth}
    \includegraphics[width=\textwidth]{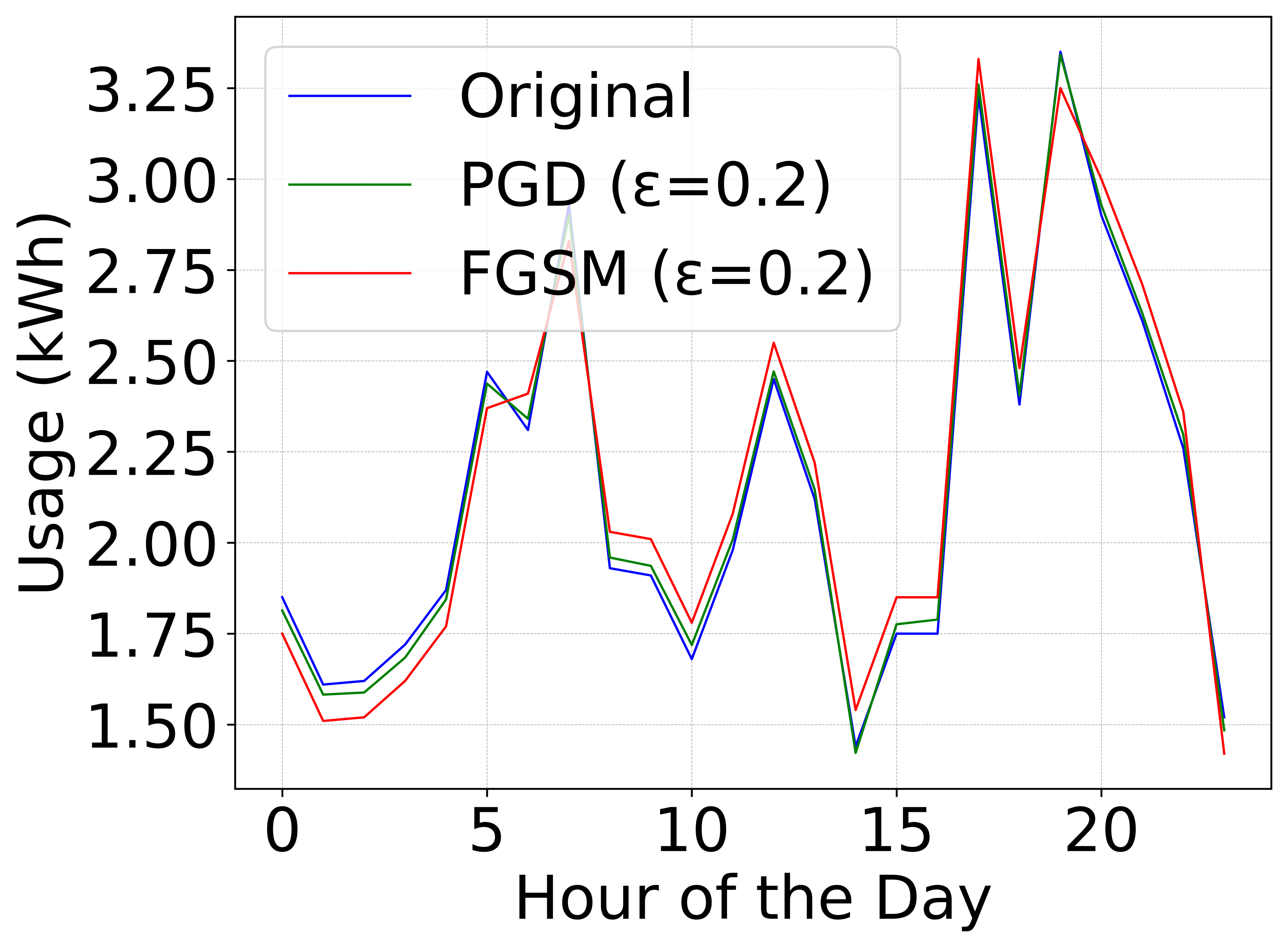}
    \caption{$\epsilon=0.2$}
    \label{fig:adv_sampless_eps_0.2}
\end{subfigure}
\begin{subfigure}{0.48\columnwidth}
    \includegraphics[width=\textwidth]{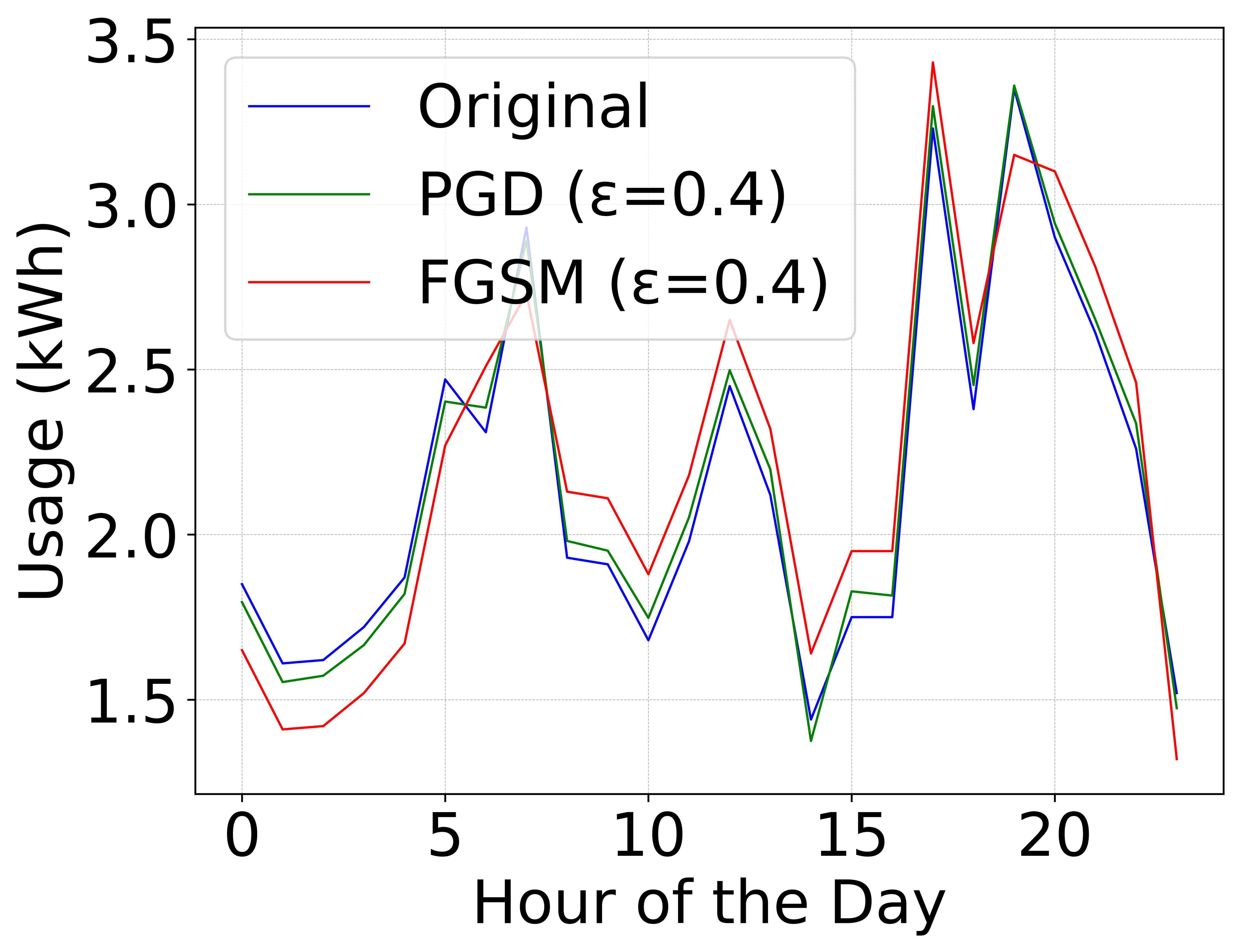}
    \caption{$\epsilon=0.4$}
    \label{fig:adv_sampless_eps_0.4}
\end{subfigure}
\begin{subfigure}{0.48\columnwidth}
    \includegraphics[width=\textwidth]{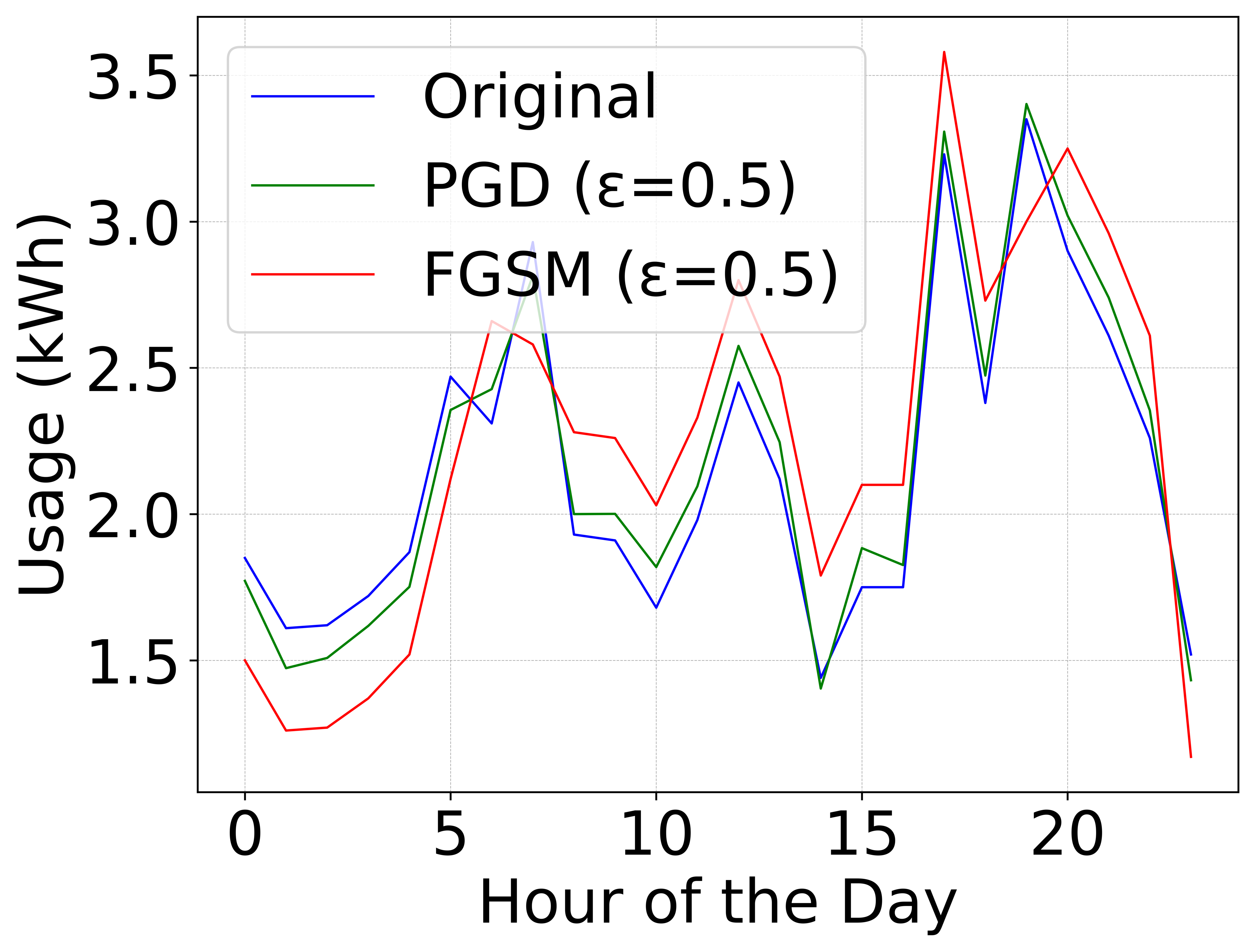}
    \caption{$\epsilon=0.5$}
    \label{fig:adv_sampless_eps_0.5}
\end{subfigure}
\begin{subfigure}{0.48\columnwidth}
    \includegraphics[width=\textwidth]{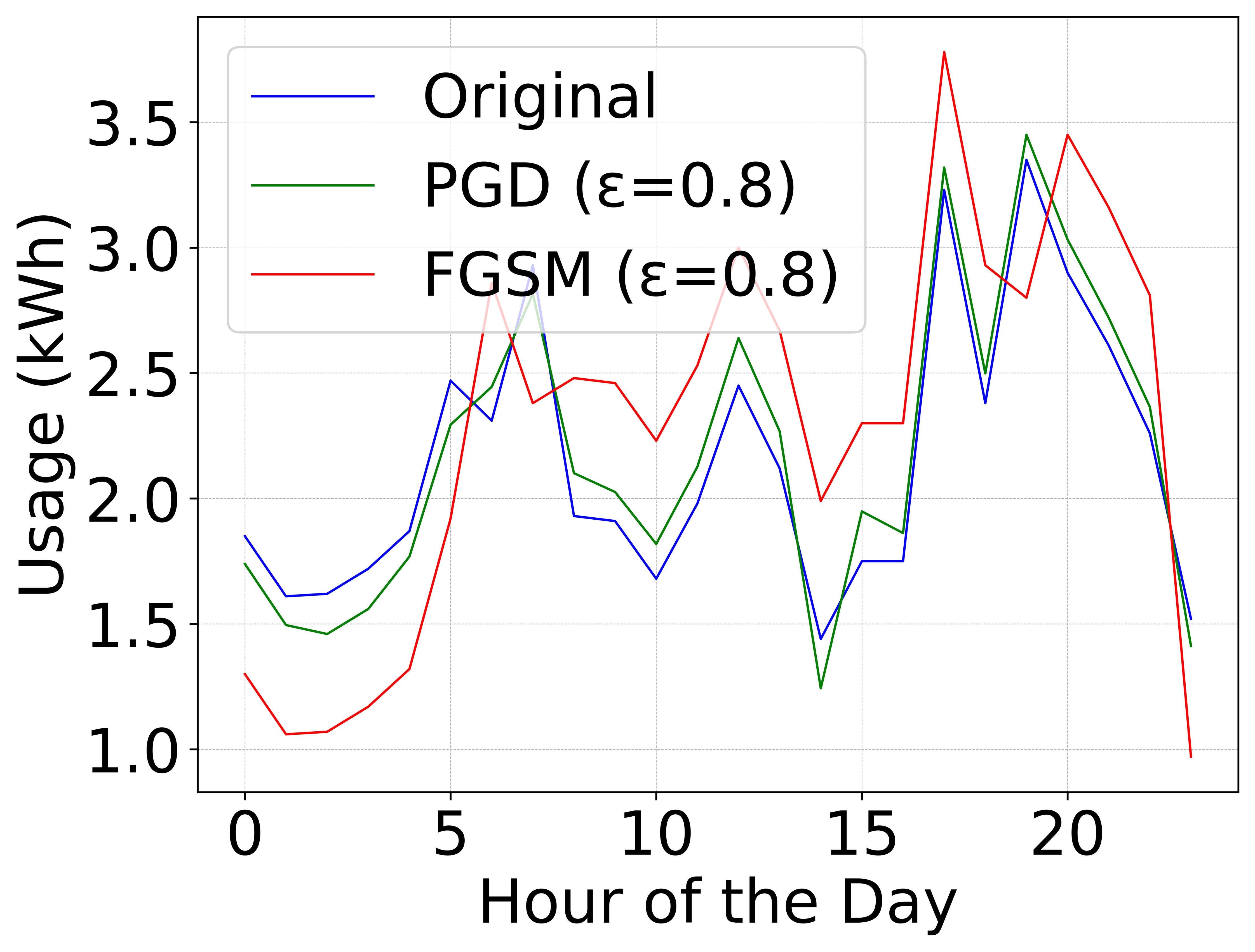}
    \caption{$\epsilon=0.8$}
    \label{fig:adv_sampless_eps_0.8}
\end{subfigure}
\caption{Clean (without attack) sample and its PGD and FGSM attacked variations with varying attack magnitudes.}
\label{fig:adv_samples}
\end{figure}

The white box attacks on the prepared time series data were generated as described in Sections \ref{sec:background} and \ref{sub_sec:AttackingAnomalyData}. Fig. \ref{fig:adv_samples} illustrates a time series sample from the dataset before attacks (clean sample), and samples after FGSM and PGD attacks with varying magnitudes of attack $\epsilon$. When $\epsilon$ is set to a lower value, the perturbations introduced to the data are smaller, making the adversarial sample closer to the clean sample. As $\epsilon$ increases, the perturbations become larger, making the adversarial sample more distinct from the clean sample. The perturbed samples by the PGD attack are still visually similar to the clean sample, whereas the FGSM attack introduces more noticeable deviations from the clean sample. The iterative process, in our experiment 10 iterations, allows the PGD attack to carefully adjust the input data for a more effective attack than FGSM.

\subsection{Model Training and Evaluation Metrics}
\label{sub_sec:ModelsandTraining}

This study examined two time series networks, LSTM and Transformer. The LSTM used in experiments comprises an LSTM layer containing 100 units, followed by a dense layer with a single unit and $\mathrm{sigmoid}$ activation to output the anomaly probability. Transformer consists of five transformer encoder blocks. Each block includes a multi-head attention layer with eight heads, each having a dimension of 160, followed by a feed-forward layer with a hidden dimension of 128. After passing through the Transformer blocks, the output across time steps is globally averaged and passed to a fully connected layer with a hidden dense layer of 256 units, followed by an output layer with a single unit. The models were trained for 100 epochs with a batch size of 32. The RMSprop optimizer was used with an initial learning rate of 0.01 multiplied by 0.1 at epochs 50, 70, and 90. Binary Focal Loss \cite{lin2017focal} was used to address the imbalance between normal and anomalous classes.

The accuracy, precision, recall, and F1 Score were used to measure the anomaly detection performance. Accuracy is the ratio of the number of correct classifications to the total number of predictions. The precision reflects the probability of a model-flagged anomaly being a ground truth anomaly, whereas the recall represents the probability of the model successfully flagging ground truth anomalies. The F1 Score is the harmonic mean of precision and recall. Measuring the success of the adversarial attacks involves comparing the model performance before and after the attack. We also calculated the Attack Success Rate (ASR), which quantifies the success probability of an adversarial attack. ASR is the ratio between the number of samples whose predicted labels are changed by the attack and the total number of samples. Higher ASR correlates with a greater degradation in model performance.
\begin{equation}
    ASR = \frac{\sum_{i=1}^N \zeta \left( y'_i \neq y_i \right)}{N}
    \label{eq:asr}
\vspace{-0.01in}
\end{equation}

\subsection{Effect of Adversarial Attacks during Inference}

\begin{table}[b]
\setlength{\tabcolsep}{3pt}
\resizebox{\columnwidth}{!}{%
\begin{tabular}{llcccc|r}     
\toprule
Setting         & Attack    & Acc(\%)       & Prec(\%)          & Rec(\%)       & F1(\%)        & ASR(\%)   \\ \midrule
LSTM            & No Attack & 95.74       & 93.91         & 93.64     & 93.78     & -         \\
(Central)       & AWGN      & 94.02       & 90.23         & 96.20     & 93.12     & 10.1    \\
                & FGSM      & 50.58       & 38.6          & 75.7      & 51.19     & 49.4    \\
                & PGD       & 18.12       & 14.4          & 31.2      & 19.7      & 86.8    \\ 
LSTM            & No Attack & 91.6        & 90.3          & 85.1      & 87.62     & -         \\
(FL)            & AWGN      & 91.36       & 92.01         & 87.38     & 89.64     & 14.2    \\
                & FGSM      & 49.79       & 37.61         & 71.84     & 48.61     & 52.98   \\
                & PGD       & 25.63       & 22.47         & 23.88     & 25.5      & 78.55   \\ \midrule 
Transformer     & No Attack & 96.24       & 94.32         & 97.37     & 95.82     & -         \\
(Central)       & AWGN      & 93.11       & 90.66         & 94.75     & 92.66     & 12.5    \\
                & FGSM      & 38.7        & 30.01         & 59.39     & 39.87     & 61.2    \\
                & PGD       & 17.9        & 21.4          & 52.4      & 30.4      & 82.03   \\ 
Transformer     & No Attack & 86.7        & 78.3          & 88.3      & 82.3      & -         \\
(FL)            & AWGN      & 86.12       & 85.12         & 90.31     & 87.64     & 17.2    \\
                & FGSM      & 34.1        & 29.1          & 33.6      & 31.2      & 71.1    \\
                & PGD       & 27.1        & 23.4          & 35.06     & 28.1      & 76.19   \\  \bottomrule
            
\end{tabular}%
}
\caption{Models trained on clean data, attacked in inference.}
\label{tab:central-model-performance}
\end{table}

In our first experiment, both federated and centralized models were trained with clean data (non-perturbed) and evaluated on the test data that contained adversarial samples. All samples in the test set were perturbed with the considered attack: FGSM, PGD, or Additive White Gaussian Noise (AWGN). An attack strength of $\epsilon=0.5$ was used for FGSM and PGD, while a noise variance of $\sigma^2 = 0.1$ was used for AWGN. Table \ref{tab:central-model-performance} illustrates the results of this experiment. The first row in each segment shows the model performance on the clean test set, and the subsequent rows show the model performance with perturbed test sets. The ASR measures the percentage of attacks in the test data that resulted in flipped predictions. 

Both centrally and federated-trained models demonstrated strong baseline anomaly detection performance on the clean test set with centrally-trained models showing better performance. FGSM and PGD attacks during inference considerably reduced the model performance, whereas AWGN resulted in minor performance reduction. Both FGSM and PGD attacks led to a higher ASR for the FL model than for the central model. Overall, PGD posed a significant challenge for both federated and centrally trained models.

\subsection{Effect of Adversarial Attacks during Federated Training}
\label{sec:results}

\begin{table}[b]
\centering
\setlength{\tabcolsep}{3pt}
\resizebox{\columnwidth}{!}{%
\begin{tabular}{llcccc|r}     
\toprule
Setting     & Attack      & Acc(\%)       & Prec(\%)      & Rec(\%)   & F1(\%)        & ASR(\%)   \\ \midrule 
LSTM        & No Attack         & 95.74      & 93.91     & 93.64  & 93.78     & -         \\
(Central)   & AWGN              & 93.45      & 94.61     & 93.74  & 92.9      & 11.71    \\
            & Label Flip        & 42.59      & 45.16     & 42.82  & 40.72     & 55.11    \\
            & FGSM              & 33.13      & 31.13     & 31.62  & 32.14     & 76.71    \\
            & PGD               & 22.36      & 19.63     & 20.43  & 21.31     & 82.41    \\
LSTM        & No Attacks        & 91.6       & 90.3      & 85.1   & 87.62     & -         \\
(FL)        & AWGN              & 90.11      & 89.66     & 82.00  & 85.66     & 15.7    \\
            & Label Flip        & 40.66      & 44.36     & 38.49  & 41.22     & 65.1    \\
            & FGSM              & 36.6       & 37.3      & 33.1   & 35.07     & 77.7    \\
            & PGD               & 23.5       & 21.4      & 24.04  & 22.6      & 85.3    \\ \midrule 
Transformer & No Attack & 96.24       & 94.32         & 97.37     & 95.82     & -         \\
(Central)   & AWGN              & 94.12      & 92.16     & 93.34  & 94.57     & 11.21    \\
            & Label Flip        & 62.27      & 60.26     & 61.75  & 63.32     & 39.17    \\
            & FGSM              & 61.25      & 61.32     & 61.71  & 62.12     & 42.29    \\
            & PGD               & 29.21      & 27.78     & 28.28  & 28.81     & 72.91    \\

Transformer & No Attacks        & 86.7       & 78.3      & 88.3   & 82.3      & -         \\
(FL)        & AWGN              & 84.61      & 80.66     & 78.68  & 79.66     & 25.1    \\
            & Label Flip        & 75.66      & 62.36     & 85.50  & 72.12     & 40.7    \\
            & FGSM              & 65.8       & 64.02     & 58.2   & 60.09     & 48.9    \\
            & PGD               & 33.01         & 29.14     & 36.21  & 32.1      & 71.7    \\ \bottomrule 
\end{tabular}%
 }
\caption{Models trained under adversarial attacks}
\label{tab:fgsm-pgd-attack}
\end{table}

In the second experiment, the adversarial attacks were introduced during training, and the anomaly detection performance was evaluated on clean test data. Attacks in the FL setting were carried on as explained in Section \ref{sub_sec:AttackingFL} where 9 out of the 19 clients were malicious, with 30\% of their training data being perturbed. The tested attack techniques consist of sophisticated attacks, PGD and FGSM, as well as naive attacks, AWGN, and label flipping. The PGD and FGSM were conducted with $\epsilon=0.5$, whereas the AWGN used $\sigma^2 = 0.1$. The central training was similarly subjected to attacks, with 30\% of the total data being perturbed by the corresponding attack during training. Table \ref{tab:fgsm-pgd-attack} shows the results of this study, where the first row in each segment shows the clean-trained model performance, and the subsequent rows show model performance after training under adversarial attacks. The ASR reflects the percentage of clean test data whose model predictions were flipped after training with attacks.

When trained in the FL setting without adversarial attacks, LSTM achieved an accuracy of 91.6\%, a precision of 90.3, a recall of 85.1, and an F1 score of 87.62 on the clean test set. Transformer, on the other hand, showed slightly lower performance with an accuracy of 86.7\%, precision of 78.3, recall of 88.3, and an F1 score of 82.3. When attacked during training, both models experienced a considerable drop in clean test set performance. ASR was high for training under FGSM, while PGD caused an even larger performance decline. Random label flipping introduced a smaller drop in performance than FGSM and PGD, while AWGN had the smallest impact on the model performance. A large performance drop under PGD attacks compared to FGSM is expected, as the iterative nature of PGD makes it a more sophisticated attack. Nevertheless, both attacks were highly successful. This is consistent with the decline observed across all performance metrics. 

Centralized training showed better performance than federated learning in the absence of attacks, which is expected since the performance of federated learning is inherently bounded by that of a centrally trained model. However, central training showed similar performance drops with attacks during training. Both LSTM and Transformer experienced substantial drops in performance when subjected to FGSM and PGD during training, with lesser impact from label flipping and the least impact from AWGN. Central training generally showed lower attack success rates than their federated-trained counterparts. Overall, the FL setting was observed to be equally or more vulnerable to all attacks than the central setting. 

Transformer demonstrated better resilience against both attacks than the LSTM in central and federated settings. The FGSM attack on Transformer had a lower success rate than on the LSTM. Similarly, although the PGD attack had a higher success rate against Transformer compared to FGSM, it was still less effective than when applied to the LSTM. Transformer exhibited a similar phenomenon for label flipping while showing higher vulnerability to AWGN than LSTM.

Additional experiments were conducted on a larger residential dataset of 50 households, also provided by London Hydro. The results, presented in Table \ref{tab:fgsm-pgd-attack-50}, reinforce the findings from the first dataset, revealing performance degradation under attacks. Both LSTM and Transformer experienced notable performance degradation under FGSM and PGD attacks during training, with PGD causing a more substantial impact. Transformer again demonstrated slight resilience compared to the LSTM, though both models were highly vulnerable.

Overall, the FL setting showed comparable vulnerability to the attacks as central learning. The FGSM and PGD attacks caused considerable performance degradation, with PGD causing the highest impact, followed by FSGM, label flipping, and random perturbations. In general, Transformer was less vulnerable to the considered attacks than the LSTM. The results emphasize the importance of considering adversarial attacks in FL, underscoring the necessity for a deeper understanding of these threats in the FL setting.

\begin{table}[t]
\centering
\setlength{\tabcolsep}{3pt}
\resizebox{\columnwidth}{!}{%
\begin{tabular}{llcccc|r}    
\toprule
Setting     & Attack          & Acc(\%)       & Prec(\%)      & Rec(\%)   & F1(\%)        & ASR(\%)   \\ \midrule
LSTM        & No Attack       & 82.3          & 82.5          & 80.3      & 81.4          & -         \\
(Central)   & AWGN            & 80.2          & 87.4          & 77.4      & 82.1          & 21.3      \\
            & Label Flip      & 59.1          & 53.7          & 61.7      & 57.4          & 45.2      \\
            & FGSM            & 35.2          & 42.1          & 32.5      & 36.7          & 61.7      \\
            & PGD             & 27.5          & 33.3          & 24.3      & 28.1          & 66.4      \\
LSTM        & No Attack       & 78.2          & 73.1          & 77.4      & 75.2          & -         \\
(FL)        & AWGN            & 78.1          & 84.3          & 74.5      & 79.1          & 25.1      \\
            & Label Flip      & 57.2          & 52.9          & 58.1      & 55.4          & 55.6      \\
            & FGSM            & 43.2          & 40.0          & 45.3      & 42.5          & 62.5      \\
            & PGD             & 33.6          & 28.6          & 35.1      & 31.5          & 68.3      \\ \midrule
Transformer & No Attack       & 84.1          & 79.6          & 85.4      & 82.4          & -         \\
(Central)   & AWGN            & 82.2          & 87.4          & 80.7      & 83.9          & 20.5      \\
            & Label Flip      & 65.9          & 61.0          & 66.2      & 63.5          & 37.8      \\
            & FGSM            & 46.1          & 50.1          & 45.3      & 47.6          & 53.1      \\
            & PGD             & 32.5          & 29.4          & 34.2      & 31.6          & 61.6      \\
Transformer & No Attack       & 74.4          & 79.3          & 72.6      & 75.8          & -         \\
(FL)        & AWGN            & 71.6          & 76.4          & 69.5      & 72.8          & 28.2      \\
            & Label Flip      & 62.2          & 66.5          & 61.1      & 63.7          & 48.6      \\
            & FGSM            & 51.4          & 46.6          & 52.6      & 49.4          & 55.4      \\
            & PGD             & 42.7          & 39.3          & 41.3      & 40.3          & 64.2      \\ \bottomrule
\end{tabular}%
}
\caption{50-houses dataset: Trained under attacks.}
\label{tab:fgsm-pgd-attack-50}
\end{table}

\subsection{Impact of Attack Magnitude on Model Performance}

This section investigates the effect of the magnitude of attacks on anomaly detection performance in the FL setting. We consider two approaches to vary the magnitude of the attack: changing the number of malicious clients in the FL setting and changing the attack strength $\epsilon$ of FGSM and PGD. 

\begin{figure}[t]
    \centering
    \begin{subfigure}[b]{0.49\linewidth}
        \centering
        \includegraphics[width=\textwidth]{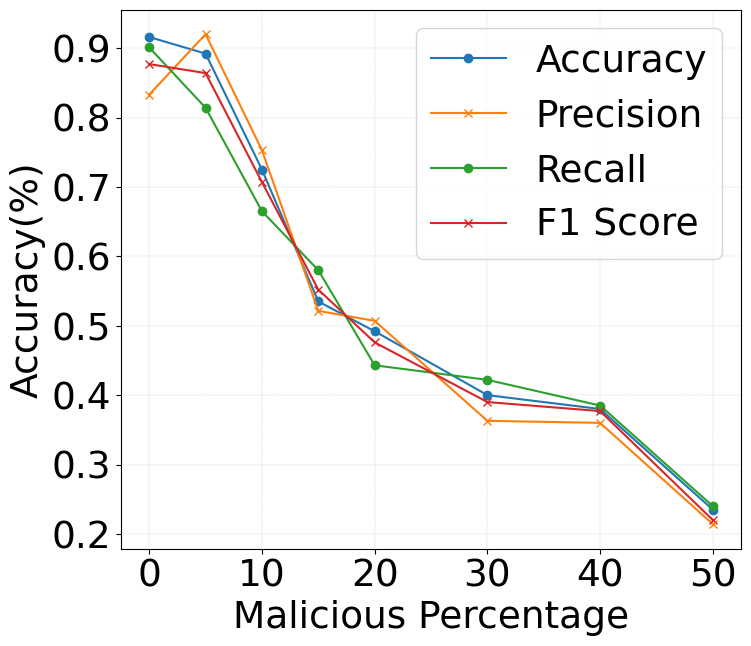}
        \caption{Acc. with malicious clients \%}
        \label{fig:malicious-percentage-acc}
    \end{subfigure}
    \hfill
    \begin{subfigure}[b]{0.49\linewidth}
        \centering
        \includegraphics[width=\textwidth]{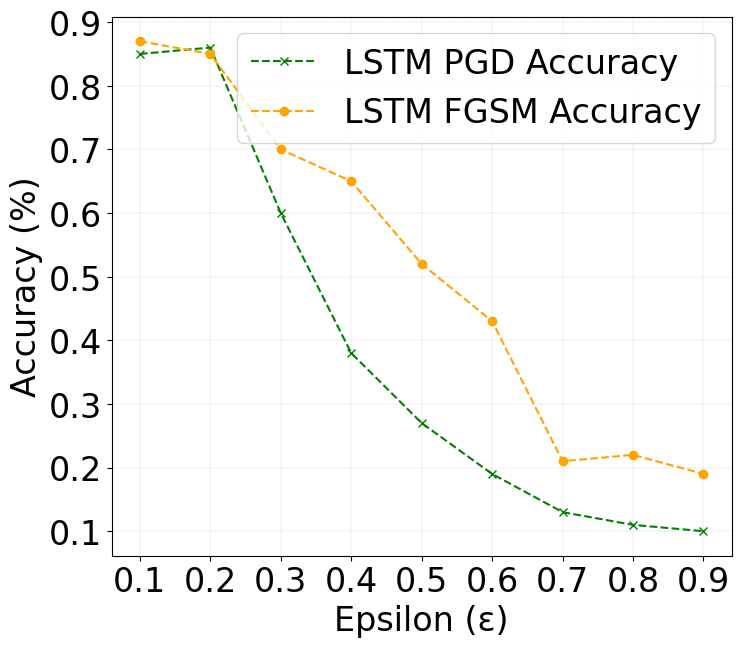}
        \caption{Acc. with attack strength \( \epsilon \).}
        \label{fig:epsilon-vs-acc}
    \end{subfigure}
    \caption{Accuracy for varied attach strengths.}
    \label{fig:attack_strength}
\end{figure}

Fig. \ref{fig:malicious-percentage-acc} illustrates the effect of the number of malicious clients performing PGD attacks on the anomaly detection performance of the LSTM in the FL setting. With 10\% malicious clients, the accuracy drops to 72.5\%. As the percentage of malicious clients increases to 20\%, the accuracy further decreases to 49.2\%. When 50\% of the clients are malicious, the model's performance is substantially compromised, with the accuracy plummeting to 23.5\%. As expected, the increasing number of malicious clients in FL results in performance degradation in terms of all considered metrics. 
Fig. \ref{fig:epsilon-vs-acc} depicts the anomaly detection accuracy of LSTM trained in the FL setting with increasing attack strength $\epsilon$ for PGD and FGSM. The parameter $\epsilon$ controls the magnitude of the perturbations applied during the attacks, with higher values resulting in larger perturbations and typically more successful attacks. For both PGD and FGSM attacks, the model's accuracy remained relatively high (around 89\%) at low $\epsilon$ values of 0.1 and 0.2, indicating that minor perturbations had minimal impact on the model's performance. However, as $\epsilon$ increased, the accuracy exhibited a large decline, with PGD causing more severe degradation than FGSM. 

These findings indicate that both PGD and FGSM attacks become increasingly effective in degrading the model's performance as the $\epsilon$ value increases. The PGD attack, being an iterative and more powerful method, had a greater effect on reducing accuracy compared to the FGSM attack.

\section{Conclusion}
\label{sec:conclusion}\vspace{-0.02in}
This paper assessed the vulnerability of FL-based anomaly detection in energy data to adversarial attacks. Two state-of-the-art deep learning models, LSTM and Transformer, were examined for their sensitivity to attacks, FGSM, and PGD, where both attacks greatly degraded model performance compared to naive attacks. PGD, being an iterative attack, consistently had a larger effect than the FGSM attack. Comparing the effect of attacks on centralized training to that of FL, the experiments show that the FL models are more affected. Transformer was less sensitive to the attacks than the LSTM in general. The strength of the attack, as well as the number of malicious clients also had a large impact on the attack's success. Overall, the results highlighted the impact of adversarial attacks on FL and the need to design robust defense mechanisms to mitigate their impact. These vulnerabilities in the energy sector can have major real-world impacts as undetected anomalies can trigger financial and safety risks. Future work will design defense mechanisms to overcome these vulnerabilities and techniques for defending against malicious clients in the FL setting.
\vspace{5pt}

\noindent\textbf{Acknowledgments:}
This work was supported by NSERC under grant ALLRP 577133-2022. Computation was enabled in part by Digital Research Alliance of Canada.

\balance
\bibliographystyle{IEEEtran}
\bibliography{references}

\end{document}